\documentclass[10pt,twocolumn,letterpaper]{article}

\usepackage{cvpr}
\usepackage{times}
\usepackage{epsfig}
\usepackage{graphicx}
\usepackage{amsmath}
\usepackage{amssymb}
\usepackage{nccmath}

\usepackage{tikz}
\usetikzlibrary{shapes.geometric}
\usetikzlibrary{positioning}
\usetikzlibrary{fadings}
\usetikzlibrary{calc}

\usepackage{color}
\usepackage{pgfplots}
\usepackage{pgf-umlsd}
\usepackage{ifthen}
\usepackage{calc}

\usepackage[font=small,labelfont=bf,tableposition=top]{caption}

\usepackage{cuted}
\usepackage{capt-of}
\usepackage{array}

\usepackage[normalem]{ulem}

\usepackage[export]{adjustbox}

\usepackage[pagebackref=true,breaklinks=true,letterpaper=true,colorlinks,bookmarks=false]{hyperref}

\newcommand{\Ltotal}[0] {{L^{total}}}
\newcommand{\Llocal}[0] {{L^{local}}}
\newcommand{\Llocalk}[0] {{L_k^{local}}}
\newcommand{\lamlocal}[0] {{\lambda_{local}}}
\newcommand{\Lda}[0] {{L_{domain}^{adv}}}
\newcommand{\lamda}[0] {{\lambda_{domain}^{adv}}}
\newcommand{\Lkl}[0] {{L^{KL}}}
\newcommand{\Lkli}[0] {{L_i^{KL}}}
\newcommand{\Lklj}[0] {{L_j^{KL}}}
\newcommand{\Lho}[0] {{L^{ho}}}
\newcommand{\Lsm}[0] {{L^{smooth}}}
\newcommand{\lamkl}[0] {{\lambda_{KL}}}
\newcommand{\lamho}[0] {{\lambda_{ho}}}
\newcommand{\lamsm}[0] {{\lambda_{smooth}}}
\newcommand{\Lr}[0] {{L^{recon}}}

\newcommand{\lamr}[0] {{\lambda_{recon}}}

\cvprfinalcopy 

\def\cvprPaperID{5321} 

\ifcvprfinal\fi
\begin{document}

\makeatletter
\def\blfootnote{\xdef\@thefnmark{}\@footnotetext}
\makeatother

\title{LADN: Local Adversarial Disentangling Network for\\  Facial Makeup and De-Makeup}

\author{
Qiao Gu$^*$\\
CMU, HKUST\\
{\tt\small qiaog@andrew.cmu.edu}
\and
Guanzhi Wang$^*$\\
Stanford University, HKUST\\
{\tt\small guanzhi@stanford.edu}
\and
Mang Tik Chiu\\
UIUC\\
{\tt\small mtchiu2@illinois.edu}
\and
Yu-Wing Tai\\
Tencent\\
{\tt\small yuwingtai@tencent.com}
\and
Chi-Keung Tang\\
HKUST\\
{\tt\small cktang@cs.ust.hk}
}


\maketitle



\thispagestyle{empty}

\setlength{\abovecaptionskip}{-5pt} 

\blfootnote{$^*$Equal contribution. Authorship order was determined by rolling dice.}
\blfootnote{This research is supported in part by Tencent and the Research
Grant Council of the Hong Kong SAR under grant no. 1620818.}

\begin{abstract}
\vspace{-10pt}
We propose a local adversarial disentangling network (LADN) for facial makeup and de-makeup. Central to our method are multiple and overlapping local adversarial discriminators in a content-style disentangling network for achieving local detail transfer between facial images, with the use of asymmetric loss functions for dramatic makeup styles with high-frequency details. Existing techniques do not demonstrate or fail to transfer high-frequency details in a global adversarial setting, or train a single local discriminator only to ensure image structure consistency and thus work only for relatively simple styles. Unlike others, our proposed local adversarial discriminators can distinguish whether the generated local image details are consistent with the corresponding regions in the given reference image in cross-image style transfer in an unsupervised setting. Incorporating these technical contributions, we achieve not only state-of-the-art results on conventional styles but also novel results involving complex and dramatic styles with high-frequency details covering large areas across multiple facial features. A carefully designed dataset of unpaired before and after makeup images is released at \href{https://georgegu1997.github.io/LADN-project-page}{https://georgegu1997.github.io/LADN-project-page}.
\vspace{-10pt}
\end{abstract}

\section{Introduction}

\begin{figure}[t!]
\begin{center}
\input{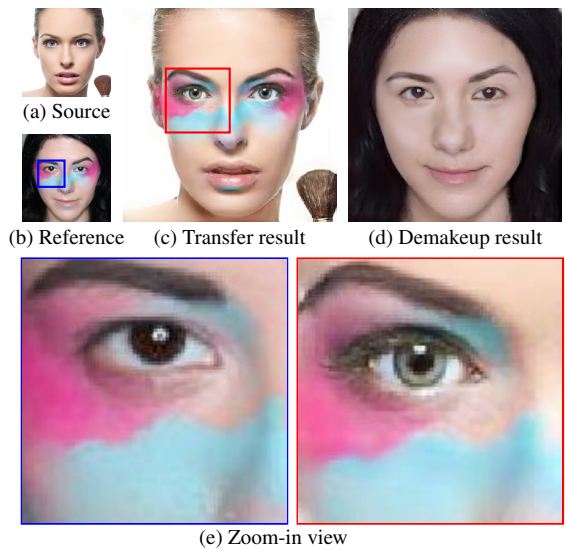}
\vspace{-.1in}
\end{center}
  \caption{Facial makeup and de-makeup with dramatic makeup style.  See supplemental material for high-quality images of all our results.}
\label{fig:teaser}
\vspace{-.15in}
\end{figure}

We propose to incorporate local adversarial discriminators into an image domain translation network for details transfer between two images, and apply these local adversarial discriminators on overlapping image regions to achieve image-based facial makeup and removal. By encouraging cross-cycle consistency between input and output, we can disentangle the makeup latent variable from other factors on a single facial image. Through increasing the number and overlapping local discriminators, complex makeup styles with high-frequency details can be seamlessly transferred or removed while facial identity and structure are both preserved. See Figure~\ref{fig:teaser}.

The contributions of our paper are:
\begin{itemize}
\item By utilizing {\em local adversarial discriminators} rather than cropping the image into different local paths, our network can seamlessly transfer and remove dramatic makeup styles;
\vspace{-0.1in}
\item Through incorporating {\em asymmetric loss functions} on makeup transfer and removal branches, the network is forced to disentangle the makeup latent variable from others, and thus our network can generate photo-realistic results where facial identity is mostly preserved;
\vspace{-0.1in}
\item A dataset containing unpaired before-makeup and after-makeup facial images will be released for non-commercial purpose upon the paper's acceptance.
\end{itemize}

Our target application, digital facial makeup~\cite{taaz,meitu}, has been increasingly popular. Its inverse application, known as facial de-makeup~\cite{Wang_2016_facebehind, makeupgo} is also starting to gain more attention. 
All current results in deep learning only either work for or demonstrate conventional or relatively simple makeup styles, possibly due to limitations of their network architectures and overfitting to their datasets. Existing methods often fail to transfer/remove dramatic makeup, which oftentimes is the main usage for such an application, before the user physically applies the dramatic makeup which may take hours to accomplish.

Given an image of a clean face without makeup as the source, and another image of an after-makeup face as the reference, the makeup transfer problem is to synthesize a new image where the specific makeup style from the reference is applied on the face of the source (Figure~\ref{fig:teaser}). 
The main problem stems from the difficulty of extracting the makeup-only latent variable, which is required to be disentangled from other factors in a given facial image.  
This problem is often referred to as content-style separation. Most existing works addressed this problem through region-specific style transfer and rendering~\cite{Liu_2016_IJCAI, Chang_2018_CVPR,
Liu_2013_ACMMM, Nguyen_2017_SmartM, Alashkar_2017_AAAI}. This approach can precisely extract the makeup style in specific and well-defined facial regions such as eyes and mouth where makeup is normally applied, but it limits the application range in the vicinity of these facial regions, and thus fails to transfer/remove more dramatic makeup where color and texture details can be far away from these facial features.

By incorporating multiple and overlapping local discriminators in a content-style disentangling network, we successfully perform transfer (resp. removal) of complex/dramatic makeup styles with all details faithfully transferred (resp. removed). 

\section{Related Work}

Given the bulk of deep learning work on photographic image synthesis, we will review related work in image translation and style transfer, and those on makeup transfer.
We will also review approaches that involve global and local discriminators and describe the differences between ours and theirs. 

\begin{table}
\begin{center}
\begin{tabular}{|c|c|}
\hline
Method & Work \\
\hline
\hline
Global discriminator & GAN \cite{gan}, pix2pix\cite{pix2pix2016}\\
\hline
Single local & Image completion\cite{IizukaSIGGRAPH2017, GFC_CVPR_2017}, \\
discriminator & PatchGAN \cite{Li2016PrecomputedRT}, CycleGAN\cite{cyclegan}\\
\hline
\textbf{Multiple overlapping}  & \textbf{LADN} (ours)\\
\textbf{local discriminators} & \\
\hline
\end{tabular}
\end{center}
\caption{Related works on local and global discriminators. Different from existing works, our paper applies multiple local discriminators in overlapping image regions.}
\vspace{-.1in}
\end{table}

\noindent {\bf Style transfer and image domain translation. \ }
Style transfer can be formulated as an image domain translation problem, 
which was first formulated by Taigman \etal~\cite{taigman_2016_unsupervised} as learning a generative function to map a sample image from a source domain to a target domain.
Isola \etal~\cite{pix2pix2016} proposed the pix2pix framework which adopted a conditional GAN to model the generative function. This method, however, requires cross-domain, paired image data for training. 
Zhu \etal~\cite{cyclegan} introduced the CycleGAN to relax this paired data requirement, by incorporating a cycle consistency loss into the generative network to generate images that satisfy the distribution of desired domain. 
Lee \etal \cite{DRIT} recently proposed a disentangled representation framework, DRIT, to diversify the outputs with unpaired training data by adding a reference image from the target domain as input. They encode images into a domain-invariant content space and another domain-specific attribute space. By disentangling content and attribute, the generated output adopts the content of an image in another domain while preserving the attributes of its own domain. 
However, in the context of makeup/de-makeup transfer, DRIT can only be applied when the relevant makeup style transfer can be formulated into image-to-image translation. As our experiments show, this means that only light makeup styles can be handled. 

\noindent {\bf Makeup transfer and removal. \ }
Tong~\etal~\cite{Tong_2007_CVPR} first tackled this problem by solving the mapping of cosmetic contributions of color and subtle surface geometry. However, their method requires the input to be in pairs of well-aligned before-makeup and after-makeup images and thus the practicability is limited. 
Guo~\etal~\cite{Guo_2009_CVPR} proposed to decompose the source and reference images into face structure, skin detail, and color layers and then transfer information on each layer correspondingly. 
Li~\etal~\cite{Li_2015_CVPR} decomposed the image into intrinsic image layers, and used physically-based reflectance models to manipulate each layer to achieve makeup transfer. 
Recently, a number of makeup recommendation and synthesis systems have been developed
\cite{Liu_2013_ACMMM, Nguyen_2017_SmartM, Alashkar_2017_AAAI}, 
but their contribution is on makeup recommendation and the capability of makeup transfer is limited. 
As recently the style transfer problem has been successfully formulated as maximizing feature similarities in deep neural networks,
Liu~\etal~\cite{Liu_2016_IJCAI} proposed to transfer makeup style by locally applying the style transfer technique on facial components.

In addition to makeup transfer, the problem of digitally removing makeup from portraits has also gained some attention from researchers
\cite{Wang_2016_facebehind, makeupgo}.
But all of them treat makeup transfer and removal as separate problems. 
Chang~\etal~\cite{Chang_2018_CVPR} formulated the makeup transfer and removal problem as an unsupervised image domain transfer problem. They augmented the CycleGAN with a makeup reference, so that the specific makeup style of the reference image can be transferred to the non-makeup face to generate  photo-realistic results. However, since they crop out the regions of eyes and mouth and train them separately as local paths, more emphasis is given to these regions. Therefore, the makeup style on other regions (such as nose, cheeks, forehead or the overall skin tone/foundation) cannot be handled properly.
Very recently, Li~\etal~\cite{Li_2018_beautygan} also tackled the makeup transfer and removal problem together by incorporating ``makeup loss'' into the CycleGAN. Although their network structure is somewhat similar, we are the first to achieve  disentanglement of  makeup latent and transfer and removal on extreme and dramatic makeup styles. 

\noindent {\bf Global and local discriminators. \ } 
Since Goodfellow \etal \cite{gan} proposed the generative adversarial networks (GANs), many related works have employed discriminators in a global setting. In the domain translation problem, while a global discriminator can distinguish images from different domains, it can only capture global structures for a generator to learn. Local (patch) discriminators can compensate this by assuming independence between pixels separated by a patch diameter and modeling images as Markov random fields.
Li \etal \cite{Li2016PrecomputedRT} first utilized the discriminator loss for different local patches to train a generative neural network. 
Such a ``PatchGAN" structure was also used in~\cite{pix2pix2016}, where a local discriminator was incorporated with an L1 loss to encourage the generator to capture local high-frequency details. 
In image completion~\cite{IizukaSIGGRAPH2017, GFC_CVPR_2017}, a global discriminator was used to maintain global consistency of image structures, while a local discriminator was used to ensure consistency of the generated patches in the completed region with the image context. 
Azadi \etal \cite{azadi2017multi} similarly incorporated local discriminator together with a global discriminator on the font style transfer problem. 

Contrary to all previous works where only a single local discriminator is used and local patches are sampled, we incorporate multiple style discriminators specialized for different facial patches defined by facial landmarks. Therefore, our discriminators can distinguish whether the generated facial makeup style is consistent with the makeup reference, and force the generator to learn to transfer the specific makeup style from the reference. 

\begin{figure}[t!]
\begin{center}
\input{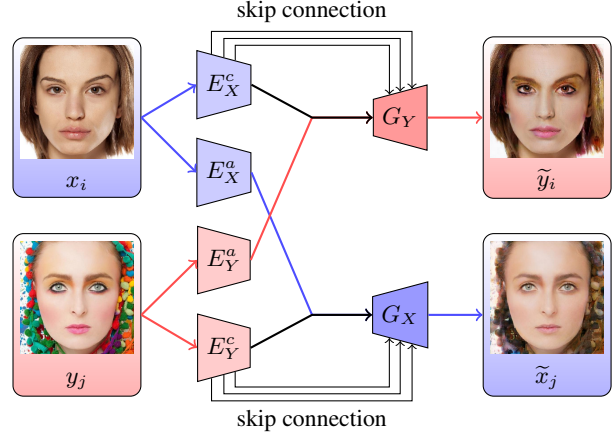}
\end{center}
  \caption{Generative Network Structure. 
  The outputs of $E^c$ and $E^a$ are $C$ and $A$, which are concatenated at the bottleneck and fed into generators. 
  Skip connections are added between $E^c$ and $G$ to capture more details in generated results. }
\label{fig:network_structure}
\vspace{-.15in}
\end{figure}

\begin{figure*}[t]
\begin{center}
\input{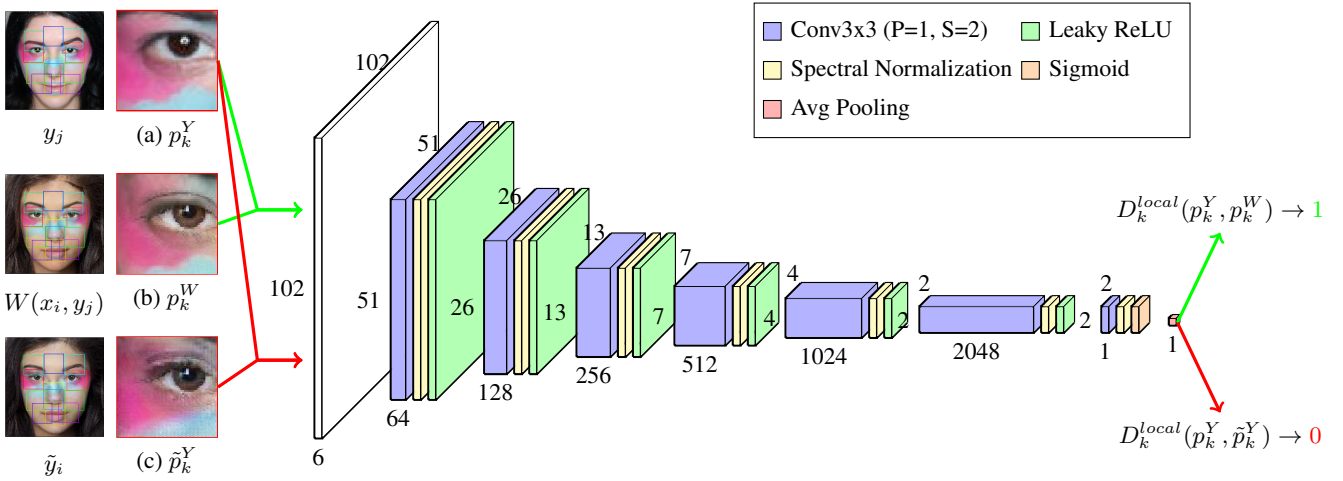}
\end{center}
  \caption{Local Patches and Local Discriminators. The local patches $p_k^Y$, $p_k^W$ and $\tilde{p}_k^Y$ (of size $102 \times 102 \times 3$) are respectively cropped from the makeup reference, the warping result and the generated image. Pairs of $p_k^Y$, $p_k^W$ are concatenated along the color channel and fed into the local discriminator as positive examples, while those of $p_k^Y$, $\tilde{p}_k^Y$ as negative ones. Each local discriminator is comprised of six $3 \times 3$ convolutional layers (padding=1, stride=2) with spectral normalization layers and leaky ReLU layers as shown. After the last layer of spectral normalization, the $2 \times 2 \times 1$ feature vector is passed to a sigmoid module and then averaged to produce a single scalar value which is the output indicating the probability of input pair possessing the same makeup style. }
\label{fig:local_patches}
\vspace{-.15in}
\end{figure*}

\section{LADN}

In the absence of adequate pixel-aligned before-makeup and after-makeup image datasets for our purpose, we will formulate makeup transfer and removal as an unsupervised image domain translation problem in section~\ref{problem_formulation}. 
We will then describe the whole network architecture and discuss our design of local style discriminators in respectively section~\ref{network_architecture} and section~\ref{local_discriminator}. Our asymmetric losses will be described in section~\ref{asymmetric_losses} with other loss functions in the network in section~\ref{loss_functions}. 

\subsection{Problem Formulation} \label{problem_formulation}

Let image domains of before-makeup faces and after-makeup faces be $X \subset \mathbb{R}^{H\times W\times 3}$ and $Y \subset \mathbb{R}^{H\times W\times 3}$ respectively.
In the unsupervised setting, we have $\{x_i\}_{i=1,\cdots,M}, x_i\in X$ to represent before-makeup examples and $\{y_j\}_{j=1,\cdots,N}, y_j\in Y$ to represent after-makeup examples, where $i$, $j$ are the identities of facial images. Note that the makeup style in $Y$ can be different for each make-up training example and there exist no before-makeup and after-makeup pairs of the same identity.  
The goal of the makeup transfer problem is to learn a mapping function $\Phi_Y: x_i, y_j \rightarrow \tilde{y}_i$, where $\tilde{y}_i$ receives the makeup style from $y_j$ while preserving the identity from $x_i$. This can be formulated as an unsupervised cross-domain image translation problem with conditioning. 
The makeup removal problem can be similarly defined as $\Phi_X: y_j \rightarrow \tilde{x}_j$, an unsupervised cross-domain image translation problem from $Y$ to $X$ without conditioning. 

\subsection{Network Architecture} \label{network_architecture}


Recently, efforts have been put on diversifying the output of cross-domain image translation. 
Latest approaches leveraging disentanglement of latent variables have shown great success in similar problems \cite{DRIT, almahairi2018augmented, huang2018munit, cao2018dida}. 
In the context of makeup transfer and removal, we want to separate the makeup style latent variable from non-makeup features (identity, facial structure, head pose, etc.) and generate new images through recombination of these latent variables. 
In this process, a disentanglement framework can suppress the false correlation between makeup style and other non-makeup features. 
Therefore, we define attribute space $A$ that captures the makeup style latent and content space $S$ which includes the non-makeup features, and our network is composed of
content encoders $\{ E^c_X, E^c_Y\}$,
style encoders $\{ E^a_X, E^a_Y\}$
and generators $\{ G_X, G_Y\}$.

As shown in Figure~\ref{fig:network_structure}, by 
$E^a_X(x_i) = A_i$, $E^a_Y(y_j) = A_j$ and
$E^c_X(x_i) = C_i$, $E^c_Y(y_i) = C_j$, 
we capture the attribute and content from a source image and a makeup reference, which are then fed into the generators to generate the de-makeup result $\tilde{x}_j$ and makeup transfer result $\tilde{y}_i$:

\begin{equation}
    G_X(A_i, C_j) = \tilde{x}_j \textrm{ and } G_Y(A_j, C_i) = \tilde{y}_i.
\end{equation}

The encoders and decoders are designed with a U-Net structure
\cite{unet},
The latent variables $A$, $C$ are concatenated at the bottleneck and skip connections are used between the content encoder and generator. This structure can help retain more identity details from the source in the generated image. For the cross-domain image adaptation, we incorporate two discriminators $\{ D_X, D_Y\}$ for the non-makeup domain and makeup domain, which tries to discriminate between generated images and real samples and thus helps the generators synthesize realistic outputs. And this gives the adversarial loss $\Lda = L_{X}^{adv} + L_{Y}^{adv}$, where
\begin{equation} \label{adv_loss}
\begin{split}
    L_{X}^{adv} &= \mathbb{E}_{x\sim P_X}[\log D_X(x)] 
    + \mathbb{E}_{\tilde{x}\sim G_X}[\log(1-D_X(\tilde{x}))] \\
    L_{Y}^{adv} &= \mathbb{E}_{y\sim P_Y}[\log D_Y(y)] 
    + \mathbb{E}_{\tilde{y}\sim G_Y}[\log(1-D_Y(\tilde{y}))].
\end{split}
\end{equation}

The discriminator $D$ tries to discriminate between generated images and real samples, while the generator $G$ tries to fool $D$ and thus can learn to adapt the generated results into the target domain.

\subsection{Local Style Discriminator} \label{local_discriminator}

We propose to use multiple overlapping local discriminators to realistically transfer makeup styles which may contain high-frequency details, and this sets ourselves apart from~\cite{Chang_2018_CVPR} which used specialized generators and global discriminators for three key regions and thus may miss makeup details that straddle outside of those regions.

To deal with the lack of ground truth of makeup transfer $y_i$, inspired by~\cite{Chang_2018_CVPR}, we generate synthetic ground truth $W(x_i, y_j)$ by warping and blending $y_j$ onto $x_i$ according to their facial landmarks. Although the synthetic results cannot serve as the real ground truth of the final results, they can provide guidance to the makeup transfer network on what the generated results should look like. Note that the warping results sometimes possess artifacts, which can be fixed by the network in the generated results. 
Based on this idea, we use local discriminators to construct style loss, which can help the generator capture the style from the makeup reference in an adversarial learning process. 
A typical placement of local discriminators is shown in Figure~\ref{fig:local_patches}, where corresponding patches in the makeup reference $y_j$, warping reference $W(x_i, y_j)$, and generated image $\tilde{y}_i$ are marked with bounding boxes. Given the image resolution of $512 \times 512$, each local discriminator considers a local image patch of size $102 \times 102$. 
Note that the local discriminators are overlapping, with one discriminator trained on one key facial landmark and thus exact locations of local discriminators are not critical. 

Given a set of $K$ local discriminators $\{D^{local}_k\}_{k=1,\cdots, K}$ at each facial landmark, a local patch from the makeup reference $p^{Y}_k$ (Figure~\ref{fig:local_patches}a), the corresponding local patch from makeup warp $p^{W}_k$ (Figure~\ref{fig:local_patches}b), and that from the generated facial image $\tilde{p}_k^Y$ (Figure~\ref{fig:local_patches}c) are cropped and fed into the local discriminator $D^{local}_k$ in pairs. By setting the ground truth for different pairs to local discriminators, the local discriminators will learn to judge $p^{Y}_k$ and $p^{W}_k$ as positive pairs (of the same makeup style), and judge $p^{Y}_k$ and $\tilde{p}_k^Y$ as negative pairs (of different makeup styles). 
Meanwhile, the goal of the generator $\Phi_Y$ is to generate the result $\tilde{y}_i$ which is of the same makeup style as the makeup reference $y_j$, therefore forming an adversarial learning process with the local discriminators. The loss for local discriminators is $\Llocal=\sum_k \Llocalk$, where $\Llocalk$ is defined as
\begin{equation} \label{local_loss}
\begin{split}
    \Llocalk
    =&\mathbb{E}_{x_i\sim P_X, y_j\sim P_{Y}}
    [\log D^{local}_k(p_k^Y, p_k^W)]\\
    +&\mathbb{E}_{x_i\sim P_X, y_j\sim P_{Y}}
    [\log [1-D^{local}_k(p_k^Y, \tilde{p}_k^Y)].
\end{split}
\end{equation}

The corresponding mini-max game is defined as
\begin{equation}
    \max_{D^{local}_k}
    \min_{E^c_X, E^a_Y, G_Y}
    \Llocal.
\end{equation}

By this setup, we take the synthetic results as guidance and encourage the local discriminators to capture makeup details from the makeup reference. Figure~\ref{fig:local_patches} gives the network details of local discriminators. 


\subsection{Asymmetric Losses}
\label{asymmetric_losses}

While transfer and removal of light makeup styles mainly involve re-coloring of eyeshadows and lips, extreme makeup style poses new challenges in this problem. 
On the one hand, extreme makeup styles contain high-frequency components, for which the network needs to differentiate from other high-frequency facial textures (e.g., eyelashes). 
On the other hand, in some cases of extreme makeup removal, the original facial color of the person can hardly be observed from the after-makeup image (e.g., Figure \ref{fig:demakeup_result} (c)), which requires the network to reconstruct or hallucinate the facial skin color without makeup. 
To tackle these challenges, 
we incorporate a high-order loss $\Lho$ for the makeup transfer branch to help transfer high-frequency details, and a smooth loss $\Lsm$ for the de-makeup branch, based on the assumption that facial colors behind the makeups are generally smooth.

{\bf High-Order Loss:} Since the warping image $W(x_i, y_j)$ preserves most texture information of the makeup style (color changes, edges) from the reference image $y_j$, we apply Laplacian filters to $p_k^W$, $\tilde{p}_k^Y$ and define high-order loss as
\begin{equation} \label{high-order_loss}
\begin{split}
    \Lho = \sum_k{h_k ||f(p_k^W)-f(\tilde{p}_k^Y)||_1},
\end{split}
\end{equation}
where $h_k$ is the weight for local patches, and $f$ is the Laplacian filter. We set $h_k$ to be similar for all local patches, with slight emphases on eye regions as eye makeups can contain subtle but essential details.

{\bf Smooth Loss:} Contrary to the makeup transfer result $\tilde{y}_i$, we do {\em not} want the de-makeup result $\tilde{x}_j$ to possess high-frequency details and instead it should be smooth in local parts. Therefore we apply a smooth loss to $\tilde{x}_j$, which is defined as
\begin{equation} \label{smooth_loss}
\begin{split}
    &\Lsm = \sum_k{s_k ||f(\tilde{p}_k^X)||_1},
\end{split}
\end{equation}
where $\tilde{p}_k^X$ is a local patch from $\tilde{x}_j$, $s_k$ is the weight for $\tilde{p}_k^X$ and $f$ is a Laplacian filter. Different from $\Lho$, we give significantly smaller weights to eyes areas since we do not want to lose the high-frequency texture around the eyes. For cheek and nose areas, we assign larger weight and thus impose a higher degree of smoothness on these regions.

The smooth loss tries to prevent the high-frequency component from presenting in $\tilde{x}_i$, while the high-order loss tries to extract and incorporate this component into $\tilde{y}_j$. Therefore, these asymmetric losses work in tandem with each other to further improve the disentanglement of the makeup latent variable from non-makeup ones. 

\subsection{Other Loss Functions} \label{loss_functions}




{\bf Reconstruction Loss:} Inspired by CycleGAN~\cite{cyclegan}, we add the reconstruction losses into the network. We feed $A_i$, $C_i$ into $G_X$ to generate $\tilde{x}^{self}_i$, and $A_j$, $C_j$ into $G_Y$ to generate $\tilde{y}^{self}_j$, which should be identical to $x_i$ and $y_j$ respectively. This gives us self reconstruction loss. 
From the generated results $\tilde{x}_j$ and $\tilde{y}_i$, we again extract the attributes and contents and use them to generate $\tilde{x}^{cross}_i$ and $\tilde{y}^{cross}_j$, which should be identical to ${x}_i$ and ${y}_j$. This gives us cross-cycle reconstruction loss.
We use L1 loss to encourage such reconstruction consistency and define the reconstruction loss $\Lr$ as
\begin{equation} \label{recon_loss}
\begin{split}
    \Lr=&||x_i-\tilde{x}^{self}_i||_1 
    + 8||x_i-\tilde{x}^{cross}_i||_1 +
    \\
    & ||y_j-\tilde{y}^{self}_j||_1 
    + 8||y_j-\tilde{y}^{cross}_j||_1,
\end{split}
\end{equation}
where we use an additional scaling factor $8$ for cross-cycle reconstruction loss to encourage the makeup transfer result to possess the makeup style. 

{\bf KL Loss:} We encourage the makeup style representation $\{A_i, A_j\}$ captured by attribute encoders $\{ E^a_X, E^a_Y\}$ to be close to a prior Gaussian distribution. Therefore, we apply KL loss $\Lkl=\Lkli+\Lklj$, where
\begin{equation} \label{kl_loss}
\begin{split}
    \Lkli=\mathbb{E}[(D_{KL}(A_i||N(0,1))],\\
    \Lklj=\mathbb{E}[(D_{KL}(A_j||N(0,1))],\\
    \textrm{ and } D_{KL}(p||q)=\int{p(x)\log{\bigg{(}\frac{p(x)}{q(x)}\bigg{)}}dx}.
\end{split}
\end{equation}

{\bf Total Loss:} Our total loss is
\begin{equation} \label{total_loss_test}
\begin{split}
    \Ltotal =& \lamlocal \Llocal + \lamda \Lda + \lamr \Lr+ \\&
    \lamkl \Lkl + \lamho \Lho + \lamsm \Lsm,
\end{split}
\end{equation}
where $\lamlocal$, $\lamda$, $\lamr$, $\lamkl$, $\lamho$, $\lamsm$ are the weights to balance different objectives. We will provide more details of setting these weights in section~\ref{training_details}.

\begin{figure}[t]
\begin{center}
\input{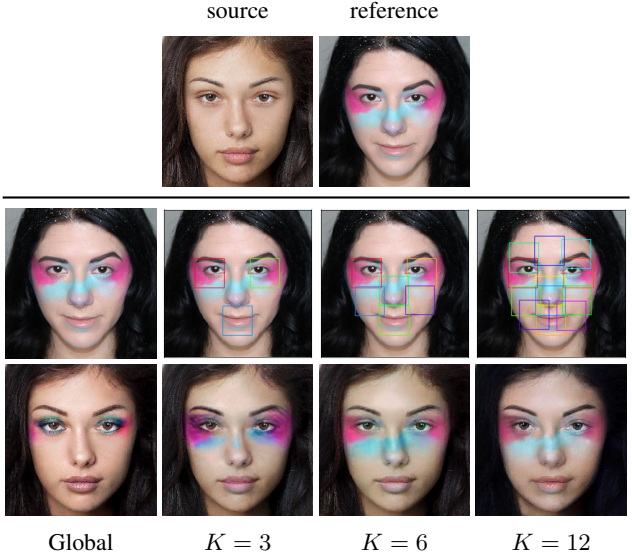}
\end{center}
  \caption{Results of global and local style discriminators. We evaluate the network outputs using only the global style discriminator (the first column) and 3, 6, 12 local style discriminators without the global one (the second, third and the fourth column respectively). First row: input source and reference images; second row: the placement of the local patches; third row: the makeup transfer results under different settings. }
\label{fig:local_D_comparison}
\vspace{-0.15in}
\end{figure}

\section{Experiments}

For results on conventional/light makeup transfer, removal and user studies for comparison with state-of-the-arts, please refer to the supplemental materials. In this section we focus on results on complex and dramatic makeups where no existing work had demonstrated significant results.

\subsection{Data Collection}

Since most datasets of face images are for recognition or identification tasks, they generally lack the labels necessary of facial makeup. There are only a few datasets on makeup that are publicly available, but most are of inadequate resolution. Some of them only contain makeup faces generated using commercial software, and thus the range of makeup styles are very limited.

As a result, we collected our own dataset, starting by collecting high-quality images of faces without occlusion from the Internet. We used facial landmark detector to filter out images without a frontal face. We then labeled a small portion of them based on the presence of makeup, from which the histogram of hue values of eyeshadow and lips regions were extracted and used to train a simple multilayer perceptron classifier. We utilized the classifier to label the remaining images and finally obtained 333 before-makeup images and 302 after-makeup images.

To achieve extreme makeup transfer, we manually selected and downloaded facial images with extreme makeup by visually inspecting whether each makeup extends out of the lip and eyeshadow regions. We obtained 115 extreme makeup images with great variance on makeup color, style and region coverage, and incorporated them into the after-makeup image set.

\begin{figure}[t]
\begin{center}
\input{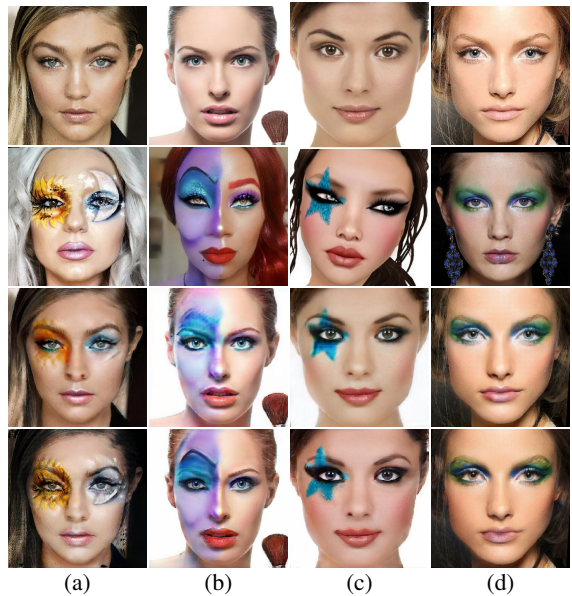}
\end{center}
  \caption{Makeup transfer results and ablation study on local high-order loss. First row: source images; second row: makeup references; third row: makeup transfer results from the network without local high-order loss; fourth row: makeup transfer results from the complete network.}
\label{fig:transfer_results}
\vspace{-0.15in}
\end{figure}


\subsection{Training Details}
\label{training_details}

We incorporate $K=12$ local discriminators into our network and set $\lamlocal=2$, $\lamda=1$, $\lamr=80$, $\lamkl=0.01$, $\lamho=20$, $\lamsm=20$. For $\Lho$, we set $h_k$ to $4$ for areas containing eyelashes, eyelids, and $2$ for areas covering nose and mouth. In $\Lsm$, $s_k$ is set to $4$ for cheek, nose areas and $0.1$ for eye areas. To balance losses while number of local discriminators varies, we additionally normalize losses from the local discriminators $\Llocal$ and losses related to local patches $\Lho$ and $\Lsm$ by $1/K=1/12$. 
The whole network is initialized by normal initialization with $\mathit{mean}=0, \mathit{gain}=0.02$. We use an Adam optimizer~\cite{kingma2014adam} with a learning rate of $0.001$ and exponential decay rates $(\beta_1,\beta_2)=(0.5,0.999)$. The resolution of input and output images is $512 \times 512$ and batch size is set to $1$ due to the GPU memory limitation. 
The network is trained firstly for 700 epochs with $\lamsm=\lamho=0$ to get stable with normal makeup styles, and then it is trained for 2000 epochs with $\lamho=20$, $\lamsm=20$ to boost performance on extreme and dramatic makeup styles. 
The input facial images are cropped and frontalized according to facial landmarks, and outputs are cropped back similarly. 

\subsection{Local Discriminators}
\label{local_discriminator_compare}

To evaluate the effect of local discriminators, comparative experiments were conducted under the settings of a single global discriminator and varying numbers of local discriminators. The last row of Figure \ref{fig:local_D_comparison} shows that the network with only a single global style discriminator fails to capture the complete makeup style from the makeup reference, only adding some random color around eyes. 
In contrast, the network becomes focused on details of makeup style when local discriminators are incorporated (As we can tell the blue and pink texture around eyes appears for $K=3$ compared to the Global case, which corresponds to the makeup reference). 
Moreover, using more local discriminators can further improve the coverage and accuracy of the transferred makeup style. As shown in Figure \ref{fig:local_D_comparison}, with the expanding coverage of local discriminators ($K=3, 6, 12$), 
the blue/red belt on the face gradually merges, while the texture on the nose shows stronger resemblance to the reference makeup, particularly in $K=12$ than $K=6$. 
Therefore, multiple and overlapping local discriminators are of paramount importance for our network to perform well, which makes feasible the transfer of complex makeup styles with high-frequency details covering large facial areas.

\begin{figure}[t]
\begin{center}
\input{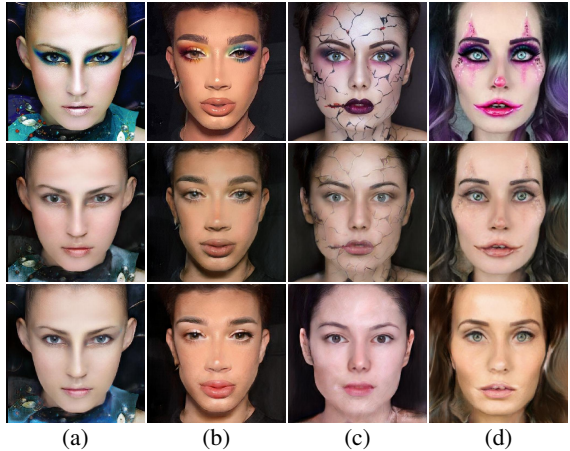}
\end{center}
  \caption{Makeup removal results and ablation study on smooth loss. First row: makeup reference; second row: de-makeup results from the network without smooth loss. third row: de-makeup results from the complete network for different styles, from (a)--(b) light and conventional style to (c)--(d) heavy and dramatic style.}
\label{fig:demakeup_result}
\vspace{-0.2in}
\end{figure}

\begin{figure*}[ht!]
\begin{center}
\input{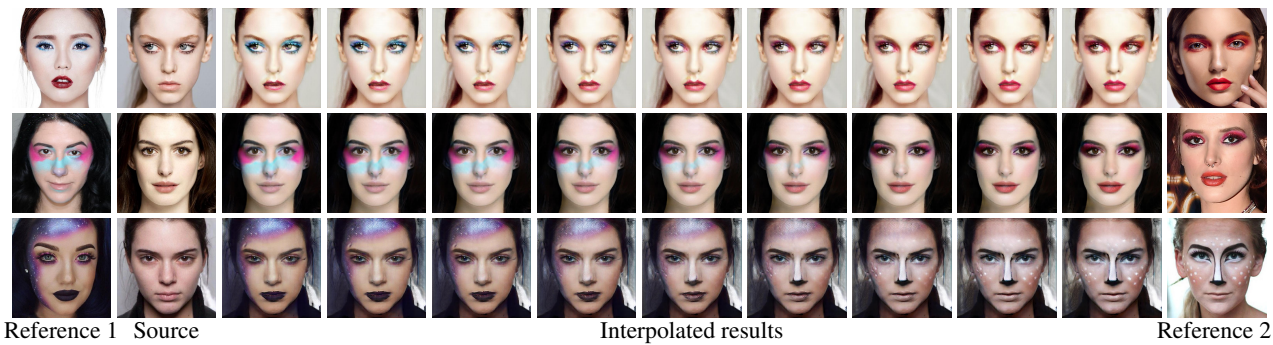}
\end{center}
\vspace{-0.1in}
  \caption{Interpolated makeup styles.  }
\label{fig:interpolation}
\vspace{-0.2in}
\end{figure*}


\subsection{Makeup Transfer Results}
As shown in Figure \ref{fig:transfer_results}, our network can transfer the makeup styles from highly dramatic ones (Figure \ref{fig:transfer_results}b) to those only on eyes and mouth (Figure \ref{fig:transfer_results}d) with considerable accuracy. 
Although the results are not perfect as the stars in Figure \ref{fig:transfer_results}a disappear in the transfer result, and the color on eyeshadows is a bit harsh, LADN is the first method to transfer and remove such dramatic makeup effects. 

To test the effect of the high-order loss on transfer results, we ran the network with the high-order loss disabled, and the results are shown in the third row of Figure~\ref{fig:transfer_results}. Comparing the third and the fourth row, we can clearly observe that some fine details get blurred in absence of the high-order loss, and this is more severe for the makeup style with more edges (for a, b and c).



\begin{figure}[ht!]
\begin{center}
\input{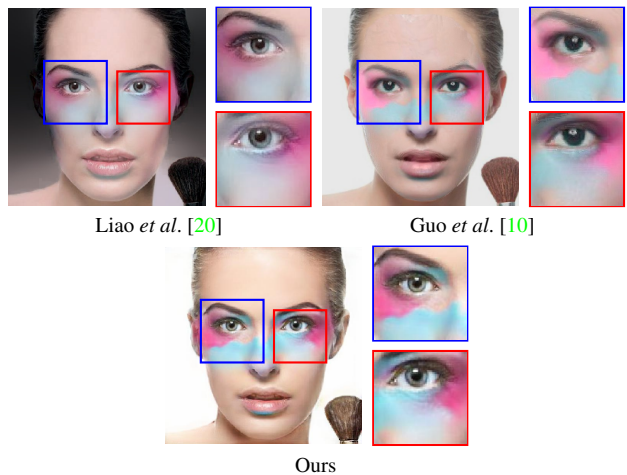}
\end{center}
\vspace{-0.05in}
  \caption{Comparison on extreme makeup transfer.}
\vspace{-0.25in}
\label{fig:extreme_comparison}
\end{figure}

\subsection{Makeup Removal Results}

Figure~\ref{fig:demakeup_result} shows makeup removal results where the styles span across a spectrum from light to heavy/dramatic makeup. As discussed in section~\ref{asymmetric_losses}, the makeup removal problem is ill-posed in the sense that there can be multiple possible faces behind the same makeup style. This can be reflected by the makeup reference in Figure~\ref{fig:demakeup_result}c, where the makeup style covers almost the whole face and the network has no clue about the face color behind the makeup from the given image.
Using the asymmetric losses, our network succeeds in distinguishing makeup style from facial texture and removing it. Then the generator hallucinates to give the identity a reasonable skin color. 

We also demonstrate the efficacy of smooth loss on extreme makeup removal by the ablation study in Figure~\ref{fig:demakeup_result}. Similar to the high-order loss, smooth loss demonstrates its significance especially when the makeup involves edges striding over large areas out of the mouth and eyes areas (for c and d). 
Meanwhile, the network can also generate satisfactory light makeup removal results (Figure~\ref{fig:demakeup_result}a and \ref{fig:demakeup_result}b). The applied eyeshadow and the lipstick are removed, recovering the normal face color without significant changes to other no-makeup areas. 


\subsection{Qualitative Comparison}

To our best knowledge, we are the first to achieve transfer and removal of dramatic makeup styles requiring no extra inputs. Figure~\ref{fig:extreme_comparison} shows a qualitative comparison on extreme makeup transfer between our results and those from~\cite{Guo_2009_CVPR} and~\cite{Liao_2017_deepimageanalogy}. 
For~\cite{Liao_2017_deepimageanalogy}, we show the result after refinement step, because otherwise the identity of the source will be lost. But as shown the refinement also blurs the makeup details. 
Similarly, the result from \cite{Guo_2009_CVPR} depicts artifacts such as discontinuity along the boundary of skin area and color fading of the makeup. Note the method in~\cite{Guo_2009_CVPR} is based on conventional methods which requires extreme accuracy on face geometry alignments between two faces. In contrast, our method only requires roughly correct landmarks in order to define the location of local discriminators.

\subsection{Interpolated Makeup Styles}
With LADN, the attribute space is disentangled well from the content space, and we can therefore easily obtain intermediate makeup styles by interpolating two attribute vectors. 
Given attribute $A_1$ and  $A_2$ respectively extracted from makeup reference $1$ and $2$, we compute $\alpha A_1 + (1-\alpha) A_2\;(\alpha \in  [0,1])$, and feed the resulting composite attribute into the generator to yield smooth transition between two reference makeup styles. Figure~\ref{fig:interpolation} shows the interpolated results from left to right which depicts a smooth and natural transition: gradual increase on the lip color and red eyeshadow as well as a gradual decrease on the bluish-pink extreme makeup, without affecting the facial identity and facial structure of the source.
Such interpolation capability enables our LADN network to not only control the amount or heaviness of the generated makeup but also to mix two makeup styles to generate a new style as shown, thus significantly broadening the range of styles through this simple mix-and-match feature provided by LADN. 

\begin{figure}[t]
\begin{center}
\input{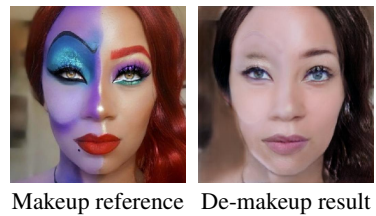}
\end{center}
\vspace{-0.05in}
  \caption{Limitation. One de-makeup example in the dataset. }
\label{fig:limitation}
\vspace{-0.25in}
\end{figure}

\vspace{-0.1in}
\section{Limitations and Conclusion}
One limitation of our network is that it struggles to remove extreme makeup styles where colors are highly consistent in local regions but vary sharply across local patches. Figure~\ref{fig:limitation} shows such style which divides the face into two halves, with each half coherent within itself (purple on left and orange on right). With very few high-frequency details present, the smooth loss is unable to take effect. As a result, the de-makeup network produces a plausible face behind the given makeup for {\em each} half of the face.
Moreover, our smooth loss is designed to encourage local smooth color transition in the de-makeup result, which is different from existing de-makeup methods which aims to recover facial imperfections before the light cosmetic makeup. Consequently, our smooth loss removes the mole as well, which in hindsight may be part of the dramatic makeup as well.



In conclusion, we propose the Local Adversarial Disentangling Network (LADN) by incorporating local style discriminators, disentangling representation and asymmetric loss functions into a cross-domain image translation network. We apply LADN to makeup transfer and removal, which demonstrates its power in transferring extreme makeup styles with high-frequency color changes and details covering large facial areas, which cannot be handled by previous work. 
Our network also achieves state-of-the-art performance in transferring and removing light/typical makeup styles. We believe this framework can also be applied to applications beyond makeup transfer and removal, which is a fruitful future research direction to explore.



{\small
\bibliographystyle{ieee_fullname}
\bibliography{main}
}

\end{document}